\documentclass{article}
\usepackage{spconf,amsmath,graphicx}

\usepackage{times}
\usepackage{latexsym}
\usepackage{caption}
\usepackage{subcaption}
\usepackage{graphicx}
\usepackage{listings}
\usepackage{flushend}

\newcommand{\asr}[0]{\textsc{asr}}
\newcommand{\wer}[0]{\textsc{wer}}
\newcommand{\sds}[0]{\textsc{sds}}
\newcommand{\iu}[0]{\textsc{iu}}

\usepackage{microtype}

\title{Evaluating Automatic Speech Recognition in an Incremental Setting}

\name{Ryan Whetten, Mir Tahsin Imtiaz, Casey Kennington}
\address{
    Boise State University\\
	Department of Computer Science\\
    1910 W University Dr., Boise, ID 83725 \\
}

\begin{document}
\maketitle
\begin{abstract}
The increasing reliability of automatic speech recognition has proliferated its everyday use. However, for research purposes, it is often unclear which model one should choose for a task, particularly if there is a requirement for speed as well as accuracy. In this paper, we systematically evaluate six speech recognizers using metrics including word-error-rate, latency, and the number of updates to already recognized words on English test data, as well as propose and compare two methods for streaming audio into recognizers for incremental recognition. We further propose Revokes per Second as a new metric for evaluating incremental recognition and demonstrate that it provides insights into overall model performance. We find that, generally, local recognizers are faster and require fewer updates than cloud-based recognizers. Finally, we find Meta's Wav2Vec model to be the fastest, and find Mozilla’s DeepSpeech model to be the most stable in its predictions.  

\end{abstract}


\section{Introduction}
Performance in automatic speech recognition (\asr) has improved dramatically in the last decade. Many \asr\ models process \emph{incrementally} in that they produce word or sub-word output as the recognition unfolds, which is an important requirement for spoken dialogue systems (\sds) that are multimodal or part of a robot platform because there is a high expectation of timely interaction from human dialogue partners \cite{kennington-etal-2020-rrsds}. Good \asr\ is critical in \sds\ applications because errors and delays produced by the \asr\ propagate to the downstream modules and overall system function. Most \asr\ models use the word-error-rate (\wer) metric to evaluate the \asr, even in conversational settings---they do not usually consider incremental metrics \cite{morris2004and}. \cite{baumann2009assessing, Baumann2016RecognisingCS} propose metrics for evaluation of incremental performance such as Edit Overhead, Word First Correct Response, Disfluency Gain, and Word Survival Rate. All of the metrics can be classified into one of the following three general areas of interest: overall accuracy, speed, and stability, but these metrics focus on discrete word-level output. 

In this paper, we make three contributions: (1) we evaluate six recent incremental \asr\ models on English data, and we also (2) propose a continuous metric that computes how much the model changes its output over time, and (3) a comparison of two methods for combining sub-word output incrementally. Following prior work \cite{morbini-etal-2013-asr,alavi-etal-2020-model,georgila-etal-2020-evaluation}, the evaluations provide for a useful guide in deciding which \asr\ model one should use. All of the models are implemented as modules in the ReTiCo framework \cite{michael2020retico} for ease of use in incremental settings.

\section{Models \& Metrics}

\begin{table*}
\small
\centering
\begin{tabular}{lll}
\hline
Name (abbreviation) & Model &  Training Data \\ \hline
Wav2Vec (W2V) & wav2vec2-base-960h & LibriSpeech \\
DeepSpeech (DS) & 0.9.3 & Fisher, LibriSpeech, Switchboard, Common Voice English \\
PocketSphinx (PS) & N/A & 1600 utterances from the RM-1\\
Vosk & en-us-0.22 & N/A \\ \hline
\end{tabular}
\caption{\label{asr-table}
Local \asr\ engines along with their used models  and training data if available.
}
\end{table*}

\begin{figure}[h]
        \centering
    \includegraphics[width=0.7\linewidth]{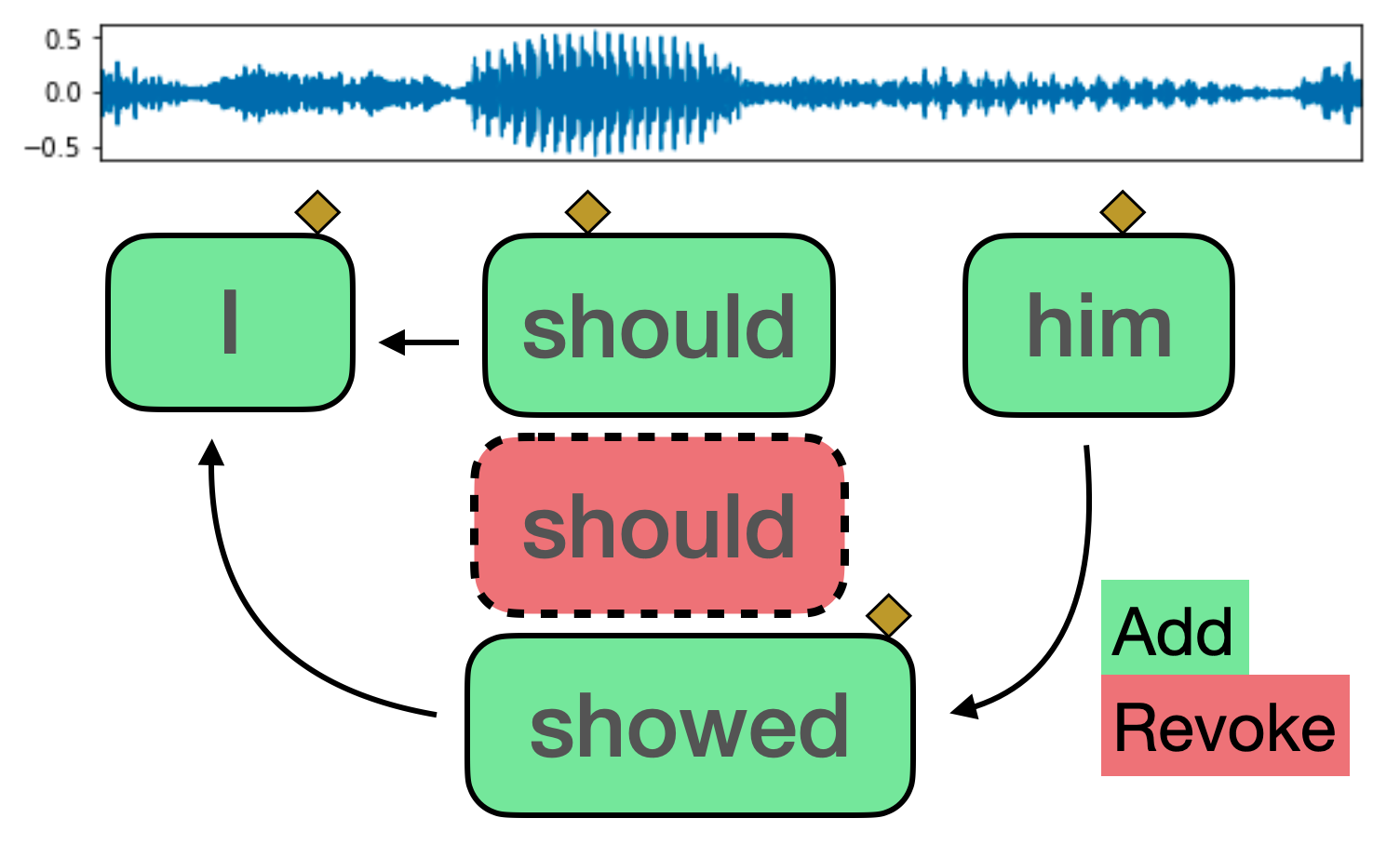}
    \caption{An example of \texttt{adds} and \texttt{revokes}. The word \emph{should} is \texttt{added}, then \texttt{revoked} and replaced by \emph{showed}. The diamonds represent the time when the predictions are made.}
    \label{fig:add-revoke}
\end{figure}

Following the evaluation strategy in \cite{Baumann2016RecognisingCS}, we adopt the \emph{Incremental Unit} (\iu) framework from \cite{schlangen-skantze-2009-general}. The \iu\ framework is practical because it is well designed and has multiple implementations from which we can build our incremental \asr\ evaluation. The framework is built around \emph{incremental units}, a discrete piece of information that is produced by a specific module. In our case, we focus on the \asr\ model as a module, and output is discretized into words (i.e., strings). The \iu\ framework has provisions for handling cases where the \asr\ output was found to be in error, given new information. The \iu\ framework proposes three operations for \iu s: \texttt{add}, \texttt{revoke}, and \texttt{commit}. A perfect \asr\ would only \texttt{add} new words to the growing recognition prefix. But as most \asr s have errors---particularly when they work incrementally---the \texttt{revoke} operation allows the \asr\ module to remove an erroneous \iu\ and replace it (i.e., through another \texttt{add} operation) in the recognized output. An example is shown in Figure \ref{fig:add-revoke}.

ReTiCo is a Python implementation of the \iu\ framework \cite{michael2020retico}. We use a ReTiCo implementation for each of the \asr\ models evaluated. We use six different, readily available \asr\ models; 2 cloud-based and 4 local (i.e., on a local GPU), chosen due to their respective results and accessibility. The cloud-based models are Google Cloud's Speech-to-Text API and Microsoft Azure's Speech SDK. Due to the limited amount of information given about the online \asr\ models, we can not go into depth about the architecture and training behind these models, but we explain the 4 local \asr\ models below. The local models are summarized in Table \ref{asr-table}.

\textbf{Wav2Vec (W2V)}: We use Meta's Wav2Vec model \cite{wav2vecpaper} from a checkpoint provided by HuggingFace where the model has been pre-trained and fine-turned on 960 hours of Librispeech \cite{baevski2020wav2vec}. This architecture is unique in that it is pre-trained on  hours of unlabeled raw audio data. While other models first convert the audio into a spectrogram, Wav2Vec operates directly on audio data.\footnote{https://huggingface.co/facebook/wav2vec2-base-960h}

\textbf{DeepSpeech (DS)}: Mozilla's DeepSpeech engine, is based on work done by \cite{hannun2014deep}. This architecture uses Recurrent Neural Networks that operate on spectrograms of the audio to make predictions. We use the 0.9.3 model and scorer for predictions. This model was trained using a wider variety of data from Fisher, LibriSpeech, Switchboard, Common Voice English, and approximately 1,700 hours of transcribed WAMU (NPR) radio shows explicitly licensed to them to be used as training corpora.\footnote{https://deepspeech.readthedocs.io/en/r0.9/}

\textbf{PocketSphinx (PS)}: One of the lighter \asr s we tested is CMU's PocketSphinx \cite{huggins2006pocketsphinx}. PS is a light-weight \asr\ that is a part of the open source speech recognition tool kit called the CMUSphinx Project. This model was trained on 1,600 utterances from the RM-1 speaker-independent training corpus. Unlike the previously mentioned models, PS does not use neural networks and is instead based on traditional methods of speech recognition by using hidden Markov models, language models, and phonetic dictionaries.\footnote{https://github.com/cmusphinx/pocketsphinx-python}

\textbf{Vosk}: Alpha Cephei's Vosk (with the vosk-model-en-us-0.22 model) is built on top of Kaldi \cite{povey2011kaldi}, and like PocketSphinx, uses an acoustic model, language model, and phonetic dictionary. Vosk uses a neural network for the acoustic model part of the engine.\footnote{https://alphacephei.com/vosk/}


\begin{figure}
\centering
        \centering
    \includegraphics[width=0.8\linewidth]{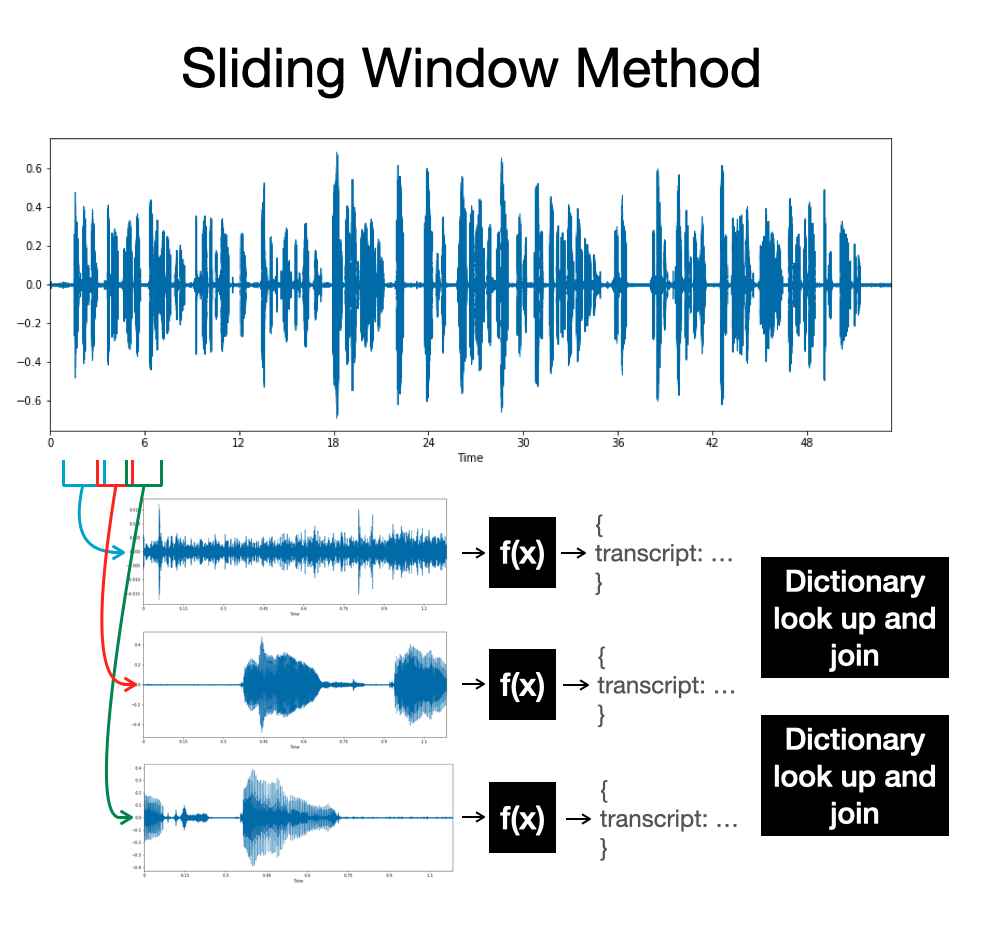}
    \caption{In the Sliding Window method, the \asr\ engine makes predictions on partially overlapping portions of audio. Dictionaries are used to join the incoming predictions together.}
    \label{fig:sliding}
\end{figure}

\subsection{Metrics}

As mentioned, all previously proposed metrics for evaluating incremental \asr\ can be divided into three broad categories: overall accuracy (using \wer), speed, and stability. We review the specific metrics used for the latter two and introduce our new metric which combines these last two categories of speed and stability into a single metric. 


\subsubsection{Predictive Speed: Latency} In order to measure the general predictive speed of an \asr\ model, we measure the time it takes from the time the \asr\ engine gets the audio until the prediction is made. We then take this time and  divide by the number of words in that particular prediction. With this, we define latency as the average amount of time per word it takes an \asr\ engine to make a prediction: $LAT= \frac{Time} {N}$, where time is measured in seconds and N is the total number words in a given prediction.

\subsubsection{Stability: Edit Overhead} For measuring stability, we measure the edit overhead (EO). EO is the total number of revokes divided by the total number of edits (additions and revokes) that the \asr\ engine makes. In an incremental \sds\ setting, this could be thought of as the fraction of text incremental units that are revokes: $  EO = \frac{R}{\#\ of\ Edits} $.

\begin{figure}
        \centering
    \includegraphics[width=0.8\linewidth]{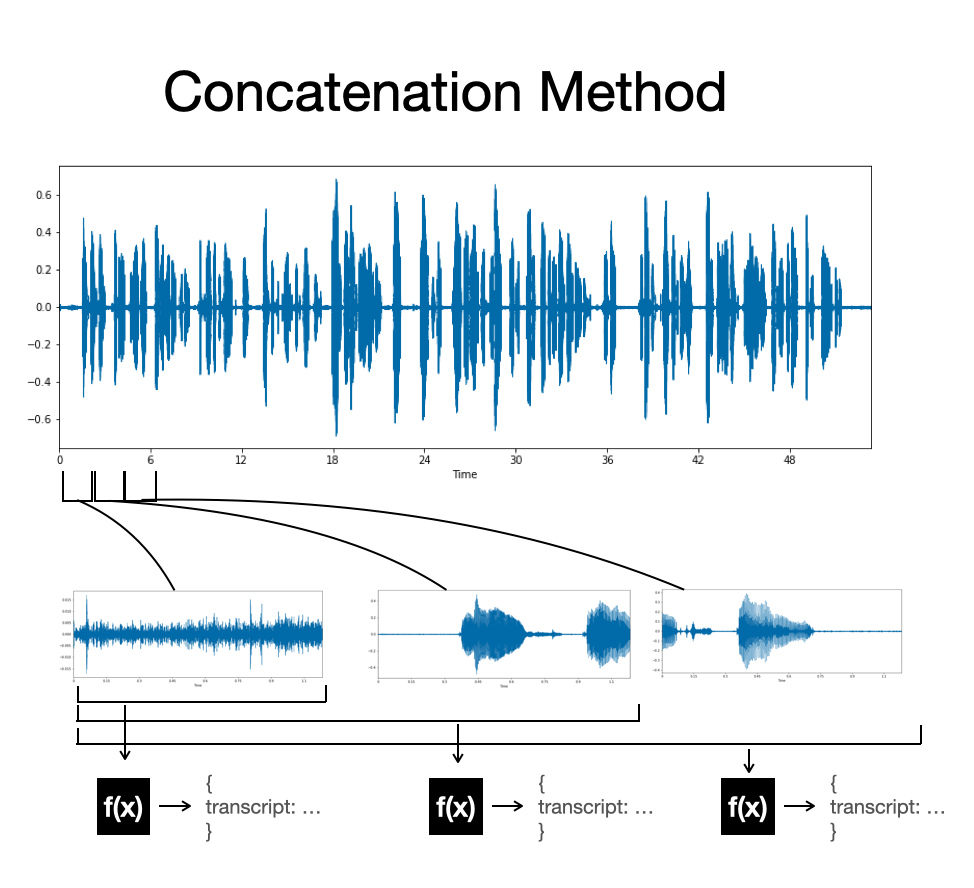}
    \caption{In this method, the incremental audio is concatenated together, and a prediction is made on the entire audio that has been given up to that point.}
    \label{fig:concat}

\end{figure}

\begin{table*}
\small
\centering
\begin{tabular}{ |c|c|c|c|c|c|c|c|c|c|c| }
\hline
\multicolumn{11}{|c|}{Incremental \asr\ Results on LibriSpeech} \\
\hline
& Google & Azure & 
W2V & W2V (Con.) & 
DS & DS (Con.) & 
PS & PS (Con.) & 
Vosk & Vosk (Con.) \\
\hline
\textbf{WER} & 13.2  & 9.1 & 10.6 & \textbf{4.0} & 18.3 & 8.4 & \emph{40.4} & 31.8 & 33.4 & 6.4 \\ 
\textbf{LAT} & 0.197 & 0.539 & \textbf{0.099} & 0.127 & 0.181 & \emph{1.443} & 0.105 & 0.220 & 0.104 & 0.167 \\ 
\textbf{EO} & \emph{0.279} & 0.065 & 0.011 & 0.093 & \textbf{0.001} & 0.013 & 0.014 & 0.147 & 0.072 & 0.019 \\
\textbf{R/Sec} & \emph{4.564} & 0.679 & 0.141 & 1.919 & \textbf{0.008} & 0.012 & 0.178 & 1.688 & 0.910 & 0.143 \\
\textbf{Sec/R} & \emph{0.219} & 1.473 & 7.083 & 0.521 & \textbf{123.135} & 80.489 & 5.613 & 0.593 & 1.099 & 7.004 \\
\hline
\end{tabular}

\begin{tabular}{ |c|c|c|c|c|c|c|c|c|c|c| }
\hline
\multicolumn{11}{|c|}{Incremental \asr\ Results on Medical Dialogue Dataset} \\
\hline
& Google & Azure & 
W2V & W2V (Con.) & 
DS & DS (Con.) & 
PS & PS (Con.) & 
Vosk & Vosk (Con.) \\
\hline
\textbf{WER} & 41.1 & \textbf{21.0} & 47.8 & 42.3 & 42.5 & 38.7 & \emph{85.6} & 80.0 & 38.4 & 23.2 \\
\textbf{LAT} & 0.287 & 0.623 & \textbf{0.125} & 0.217 & 0.245 & \emph{1.452} & 0.131 & 0.394 & 0.307 & 1.296 \\ 
\textbf{EO}  & \emph{0.243} & 0.055 & 0.016 & 0.211 & \textbf{0.000} & 0.014 & 0.005 & 0.240 & 0.048 & 0.025 \\
\textbf{R/Sec} & \emph{5.944} & 0.207 & 0.253 & 6.376 & \textbf{0.000} & 0.013 & 0.046 & 2.447 & 0.215 & 0.079 \\
\textbf{Sec/R} & \emph{0.168} & 4.837 & 3.953 & 0.157 & \textbf{inf} & 75.616 & 21.734 & 0.409 & 4.649 & 12.733 \\
\hline
\end{tabular}

\caption{\label{results-table}
Summary of results. The bold indicates the best performance and the italicized indicates the lowest performance for the given metric in the far left column. Local \asr s had lower latency than cloud-based \asr s. The Concatenation method, shown in the columns that contain a \emph{(Con.)}, had higher latency and resulted in a higher EO and RPS, but not as many revokes as the online \asr s. \emph{inf} means zero revokes per second.}
\end{table*}

\subsubsection{Revokes per Second} Our proposed and final metric is the number of \emph{Revokes per Second} (RPS). We propose this metric as way to capture the relationship between both speed and stability in an interpretable fashion. In an incremental \sds\ setting, this is the average number of \asr\ word output \iu s per second that are labeled as type \texttt{revoke}. 



We first calculate the average number of revokes per word, then divide the average number of revokes per word by our metric for latency to get the average number of Revokes per Second. We also look at the inverse \emph{Seconds per Revoke} (SPR) as a simple adjustment to this metric to see how many seconds will pass by before one can expect to see a revoke. This SPR value is useful in interpretations when the RPS is low. Taken together, the formulas for these metric are as follows:

\[ RPS = \frac{R}{N} \frac{N}{Time(s)} = \frac{R}{Time(s)} \]
\[ SPR = \frac{Time(s)}{R} = \frac{1}{RPS} \]

\subsection{Combining Sub-word Output}

Both Google and Azure  offer incremental \asr\ results. For these two \asr s, the audio files are sent to the cloud services in chunks, and the service returns a prediction with other meta-information. The local \asr\ engines work at word and sub-word levels, necessitating a method of combining the sub-word output into words.\footnote{We used the same PC with a GTX1080TI GPU for the local models.} We apply and compare two methods in this evaluation: Sliding Window and Concatenation.

For Sliding Window, we pass the audio from the file in chunks that are a bit longer than one second. These are then concatenated together as an audio buffer and given to an \asr\ model until it produces a prediction of at least 5 words or the audio buffer contains about 30 seconds of audio. At this point, we remove the first 35\% and repeat. This results in a series of predictions on segments of audio containing around 2 to 5 words. When a prediction is received, it is joined together with previous predictions. Due to overlap in incoming predictions, the way that the predictions are joined together is non-trival. The lookup method joins predictions using dictionaries from WordNet and NLTK \cite{miller1995wordnet,bird2009natural}. 

For the Concatenation method, we present the audio in chunks into an audio buffer in the same manner as the Sliding Window method, except the buffer is a concatenation of all the audio (i.e., no audio ever gets removed from the buffer). Essentially, with this method, the \asr\ model makes a prediction from the very beginning of the file to the most recent audio given to the buffer. This is computationally more expensive and takes more memory because the \asr\ model has to make predictions on longer pieces of audio as time goes on, but this method eliminates the need for joining. Diagrams showing these two methods can be seen in Figures \ref{fig:sliding} and \ref{fig:concat}.




\section{Experiment}

In this section, we explain our experiment including the evaluation data we used, and how we systematically produced and evaluated the \iu s from our \asr\ modules. 

\subsection{Data \& Procedure} \label{data}
For evaluation, we use 2 datasets, LibriSpeech and a recently assembled dialogue dataset of simulated medical conversations \cite{fareez2022dataset}.\footnote{We were unable to obtain the Switchboard corpus due to prohibitive costs.} The LibriSpeech test-clean dataset contains 5.4 hours of speech from 40 different speakers, 20 male and 20 female. This audio is divided into over 2,600 files with an average of about 20 words per file containing a vocabulary of over 8,100 words. To ensure the audio would work on all of our models, we converted the audio files to WAV files.

The medical conversation dataset contains 272 audio files with corresponding transcripts. The audio files range from around 7 to 20 minutes in length or about 800 to 2,200 words. Due to the size of these audio files, we split up the files into utterances based on silence and then randomly sample a set of 40 utterances, 17 of which were able to be processed by all 6 \asr\ engines (max 40 seconds, min 0.8 seconds, 6.1 seconds in length on average). This happened due to the length of some of the files and the constraints that each model can handle. The purpose of using this dialogue data is to 1) test each model on domain data that presumably none of them have been trained on (since this dataset was just made public in 2022), and 2) test how each engine performs on a dialogue dataset that contains disfluencies such as fillers, corrections, and restarts.









\subsection{Results}
The results can be seen in Table \ref{results-table}. When using the Sliding Window method, local models had lower latency than both the cloud models. Some of the local \asr\ models using the Concatenation method were also faster than both of the cloud ones, but generally the concatenation tests were slower and had a higher EO than the Sliding Window method. Despite this, the Concatenated versions performed better than their corresponding Sliding Window version in terms of \wer. For the cloud models, Google is less accurate and more revoke dependant than Azure. However, Google is considerably quicker which could be crucial in an interactive dialogue setting. The cloud models had surprisingly low latency (though the latency is dependent on the Internet speed), but the local \asr s generally had the lowest latency. 

The local \asr\ engine which performed the best overall in terms of \wer\ was the W2V model using the Concatenation method on the LibriSpeech data and Vosk on the Medical Dialogue data, while the model with the lowest Edit Overhead was the DS model using the Sliding Window method. Though a low \wer\ is generally better, the number of revokes has implications for downstream modules in an \sds; keeping the EO low and Revokes per Second low with a low \wer\ means the model was correct early, which is ideal.

Our results are consistent with previous evaluations on Incremental \asr\ \cite{Baumann2016RecognisingCS} that show that Google's \asr\ predictions, although fairly accurate overall, are not as stable as the others, with the highest Edit Overhead of 0.279/0.228 and an average of about 4.5/5.1 Revokes per Second on the LibriSpeech dataset and Medical Dialogue dataset respectively. 

The DS model's \wer\ was higher than other models, but the low EO and infrequent number of revokes make it a potentially good candidate for an \sds\ that requires high accuracy as well as low latency and EO, for example in a robotic platform. We suggest Concatenation for live microphones because it is more accurate and does not require a dictionary.

\section{Conclusion}
In this work, we tested six different \asr\ models in an incremental \sds\ setting and evaluated using final \wer, latency, and Edit Overhead. We also proposed a new metric, Revokes per Second. We showed that, generally, online \asr\ (Google Cloud and Azure cloud services) is not as fast as most local \asr\ engines tested, and while these are some of the most accurate \asr s we tested, they both have a relatively high number of Revokes per Second which, in combination with the latency, could potentially lead to more issues in an incremental setting. 

One of the challenges of the evaluation of \asr\ models is that, as described, the cloud \asr s do not publicly describe the precise architecture and training data used, and each of the local \asr s differs greatly in architecture and in the training data used. With this, there are too many variables and unknowns to attribute good \wer\ in a given model to its architecture, or due to training data. That being said, we do believe that in terms of testing the \emph{out of box} performance, our results are conclusive that online \asr\ tend to have higher latency and Edit Overhead. Furthermore, we also believe that our proposed metric, Revokes per Second, is an interpretable useful metric that should be used as \asr\ becomes more prevalent in \emph{live} settings such as in Spoken Dialogue Systems on a robots or in live captioning in online meetings. 

In future work, we plan on evaluating using different datasets and in different languages.   



\bibliographystyle{ieeebib}
\bibliography{iasr}

\begin{thebibliography}{10}

\bibitem{kennington-etal-2020-rrsds}
Casey Kennington, Daniele Moro, Lucas Marchand, Jake Carns, and David McNeill,
\newblock ``rr{SDS}: Towards a robot-ready spoken dialogue system,''
\newblock in {\em Proceedings of the 21th Annual Meeting of the Special
  Interest Group on Discourse and Dialogue}, 1st virtual meeting, July 2020,
  pp. 132--135, Association for Computational Linguistics.

\bibitem{morris2004and}
Andrew~Cameron Morris, Viktoria Maier, and Phil Green,
\newblock ``From wer and ril to mer and wil: improved evaluation measures for
  connected speech recognition,''
\newblock in {\em Eighth International Conference on Spoken Language
  Processing}, 2004.

\bibitem{baumann2009assessing}
Timo Baumann, Michaela Atterer, and David Schlangen,
\newblock ``Assessing and improving the performance of speech recognition for
  incremental systems,''
\newblock in {\em Proceedings of human language technologies: The 2009 annual
  conference of the north american chapter of the association for computational
  linguistics}, 2009, pp. 380--388.

\bibitem{Baumann2016RecognisingCS}
Timo Baumann, Casey~Redd Kennington, J.~Hough, and David Schlangen,
\newblock ``Recognising conversational speech: What an incremental asr should
  do for a dialogue system and how to get there,''
\newblock in {\em IWSDS}, 2016.

\bibitem{morbini-etal-2013-asr}
Fabrizio Morbini, Kartik Audhkhasi, Kenji Sagae, Ron Artstein, Do{\u{g}}an Can,
  Panayiotis Georgiou, Shri Narayanan, Anton Leuski, and David Traum,
\newblock ``Which {ASR} should {I} choose for my dialogue system?,''
\newblock in {\em Proceedings of the {SIGDIAL} 2013 Conference}, Metz, France,
  Aug. 2013, pp. 394--403, Association for Computational Linguistics.

\bibitem{alavi-etal-2020-model}
Seyed~Hossein Alavi, Anton Leuski, and David Traum,
\newblock ``Which model should we use for a real-world conversational dialogue
  system? a cross-language relevance model or a deep neural net?,''
\newblock in {\em Proceedings of the 12th Language Resources and Evaluation
  Conference}, Marseille, France, May 2020, pp. 735--742, European Language
  Resources Association.

\bibitem{georgila-etal-2020-evaluation}
Kallirroi Georgila, Anton Leuski, Volodymyr Yanov, and David Traum,
\newblock ``Evaluation of off-the-shelf speech recognizers across diverse
  dialogue domains,''
\newblock in {\em Proceedings of the 12th Language Resources and Evaluation
  Conference}, Marseille, France, May 2020, pp. 6469--6476, European Language
  Resources Association.

\bibitem{michael2020retico}
Thilo Michael,
\newblock ``Retico: An incremental framework for spoken dialogue systems,''
\newblock in {\em Proceedings of the 21th Annual Meeting of the Special
  Interest Group on Discourse and Dialogue}, 2020, pp. 49--52.

\bibitem{schlangen-skantze-2009-general}
David Schlangen and Gabriel Skantze,
\newblock ``A general, abstract model of incremental dialogue processing,''
\newblock in {\em Proceedings of the 12th Conference of the {E}uropean Chapter
  of the {ACL} ({EACL} 2009)}, Athens, Greece, Mar. 2009, pp. 710--718,
  Association for Computational Linguistics.

\bibitem{wav2vecpaper}
Alexei Baevski, Henry Zhou, Abdelrahman Mohamed, and Michael Auli,
\newblock ``wav2vec 2.0: A framework for self-supervised learning of speech
  representations,'' 2020.

\bibitem{baevski2020wav2vec}
Alexei Baevski, Yuhao Zhou, Abdelrahman Mohamed, and Michael Auli,
\newblock ``wav2vec 2.0: A framework for self-supervised learning of speech
  representations,''
\newblock {\em Advances in Neural Information Processing Systems}, vol. 33, pp.
  12449--12460, 2020.

\bibitem{hannun2014deep}
Awni Hannun, Carl Case, Jared Casper, Bryan Catanzaro, Greg Diamos, Erich
  Elsen, Ryan Prenger, Sanjeev Satheesh, Shubho Sengupta, Adam Coates, et~al.,
\newblock ``Deep speech: Scaling up end-to-end speech recognition,''
\newblock {\em arXiv preprint arXiv:1412.5567}, 2014.

\bibitem{huggins2006pocketsphinx}
David Huggins-Daines, Mohit Kumar, Arthur Chan, Alan~W Black, Mosur
  Ravishankar, and Alexander~I Rudnicky,
\newblock ``Pocketsphinx: A free, real-time continuous speech recognition
  system for hand-held devices,''
\newblock in {\em 2006 IEEE International Conference on Acoustics Speech and
  Signal Processing Proceedings}. IEEE, 2006, vol.~1, pp. I--I.

\bibitem{povey2011kaldi}
Daniel Povey, Arnab Ghoshal, Gilles Boulianne, Lukas Burget, Ondrej Glembek,
  Nagendra Goel, Mirko Hannemann, Petr Motlicek, Yanmin Qian, Petr Schwarz,
  et~al.,
\newblock ``The kaldi speech recognition toolkit,''
\newblock in {\em IEEE 2011 workshop on automatic speech recognition and
  understanding}. IEEE Signal Processing Society, 2011, number CONF.

\bibitem{miller1995wordnet}
George~A Miller,
\newblock ``Wordnet: a lexical database for english,''
\newblock {\em Communications of the ACM}, vol. 38, no. 11, pp. 39--41, 1995.

\bibitem{bird2009natural}
Steven Bird, Ewan Klein, and Edward Loper,
\newblock {\em Natural language processing with Python: analyzing text with the
  natural language toolkit},
\newblock " O'Reilly Media, Inc.", 2009.

\bibitem{fareez2022dataset}
Faiha Fareez, Tishya Parikh, Christopher Wavell, Saba Shahab, Meghan Chevalier,
  Scott Good, Isabella De~Blasi, Rafik Rhouma, Christopher McMahon, Jean-Paul
  Lam, et~al.,
\newblock ``A dataset of simulated patient-physician medical interviews with a
  focus on respiratory cases,''
\newblock {\em Scientific Data}, vol. 9, no. 1, pp. 1--7, 2022.

\end{thebibliography}


\end{document}